\documentclass{article}

\usepackage{scml2024}

\usepackage[utf8]{inputenc} 
\usepackage[T1]{fontenc}    
\usepackage{hyperref}       
\hypersetup{colorlinks,allcolors=blue}
\usepackage{url}            
\usepackage{booktabs}       
\usepackage{amsfonts}       
\usepackage{nicefrac}       
\usepackage{microtype}      
\usepackage{lipsum}         
\usepackage{graphicx}
\usepackage[numbers]{natbib}
\usepackage{doi}
\usepackage{wrapfig}
\usepackage{subfig}
\usepackage{xcolor}
\usepackage{amsmath}
\usepackage{cleveref}

\title{Discovering intrinsic multi-compartment pharmacometric models using Physics Informed Neural Networks}

\date{}

\usepackage{authblk}

\author[1,2]{%
	Imran Nasim\thanks{\texttt{imran.nasim@ibm.com}, \texttt{m20319@surrey.ac.uk}}}%
\author[2]{%
	Adam Nasim\thanks{\texttt{m20331@surrey.ac.uk}}}%
\affil[1]{IBM, UK}
\affil[2]{Department of Mathematics, University of Surrey, Guildford, GU2 7XH, Surrey, UK}

\begin{document}

\maketitle

\begin{abstract}

Pharmacometric models are pivotal across drug discovery and development, playing a decisive role in determining the progression of candidate molecules.
However, the derivation of mathematical equations governing the system is a labor-intensive trial-and-error process, often constrained by tight timelines.
In this study, we introduce PKINNs, a novel purely data-driven pharmacokinetic-informed neural network model. PKINNs efficiently discovers and models intrinsic multi-compartment-based pharmacometric structures, reliably forecasting their derivatives. The resulting models are both interpretable and explainable through Symbolic Regression methods. Our computational framework demonstrates the potential for closed-form model discovery in pharmacometric applications, addressing the labor-intensive nature of traditional model derivation. With the increasing availability of large datasets, this framework holds the potential to significantly enhance model-informed drug discovery.
\end{abstract}

\section{Introduction}
Pharmacometrics, a vital tool in drug discovery, utilizes pharmacokinetic-pharmacodynamic (PKPD) models to employ ordinary differential equations (ODEs) for describing the  relationship between dose, concentration, intensity, and response duration~\cite{Derendorf1999,Bender2015,Tuntland2014,Gibiansky2009}. These models are integral from early-phase target validation to optimizing lead compound development, scaling compounds for human dose predictions, and forecasting adverse events in toxicology. In clinical stage development, they aid in determining appropriate trial doses and predicting doses in untested populations, such as pediatric trials. Pharmacometric modeling is increasingly influential in drug submissions and clinical trial applications (CTAs)\cite{Usman2023,Rowland2015,Powell2007,Bhattaram2007,Gobburu2001}. The landscape is evolving with the rise of large datasets and the growing use of artificial intelligence (AI), with a keen interest in purely data driven approaches to model discovery.
\newline
Inverse problems widely arise across many areas in computational science \cite{iglesias2013_ensemble,sambridge2002_monte}, where the challenge frequently revolves around deducing a specific set of parameters based on certain observations or measurements. Traditional approaches based on optimization methods have shown some promise in specific areas but typically require specific domain knowledge and often have high computational costs \citep{givoli2021_adjoint,bui2013_bayes}. Physics informed neural networks (PINNs) have shown great promise in being applied to the inverse problem \citep{kharazmi2021_hp,wang2021_eigenvector,jin2021_nsfnets} by incorporating prior knowledge about the physics into the neural networks (NN). A limiting assumption used in the standard PINN studies is that the functional form of the differential equation is known, which is a particular issue for problems requiring the identification of relevant models. Very recent works have attempted to address this problem by combing the PINNs framework with symbolic regression (SR) methods \citep{chen2021_physics_SR,zhang2024_discovering_SR}. The objective of this study is to discover an underlying mathematical model based on pharmacometrics data, in a purely data driven way.
To this end, we present a novel pharamacokinetic informed neural network (PKINNs hereafter) which combines PINNs and SR, enabling the network to possess important features of interpretability and generalizability, to discover intrinsic mechanistic models from noisy data in addition to estimating several unknown parameters. 
We demonstrate that our framework accurately and robustly predicts the intrinsic derivatives of the underlying PK model and performs well in extrapolation prediction scenarios.

\section{Methodology and Experiments}

\begin{figure}[ht]
    \centering
\includegraphics[width=0.8\textwidth]{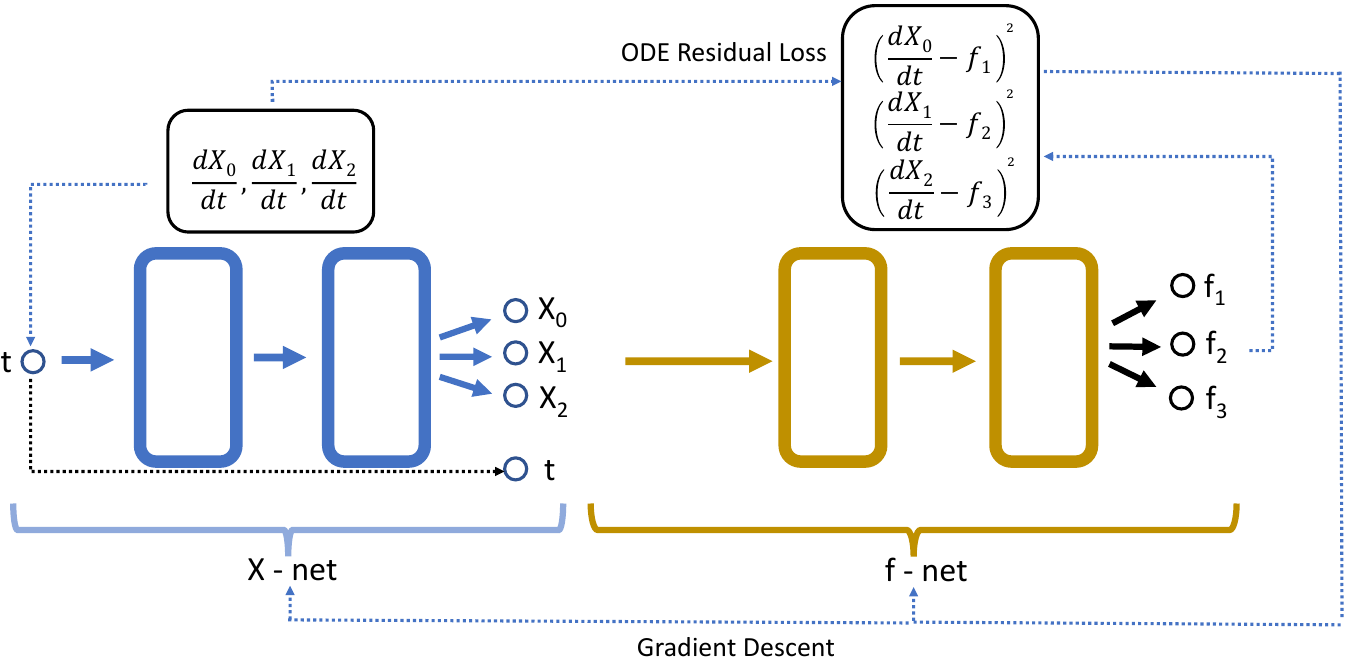}
    \caption{Schematic of the PKINNs architecture.}
    \label{fig:workflow}
\end{figure}

\textbf{Physics-informed Neural Network.}
The vanilla PINN approach \citep{raissi2019_pinns} tackles differential equation based problems by developing neural networks as substitute models for key quantities. It involves creating a loss function in relation to the differential equation that characterizes the physical phenomena adjusting the neural network parameters to minimize this loss function. The schematic of our PKINNs architecture is presented in Fig \ref{fig:workflow}. The overall network comprises of two sub-networks $X$-net and $f$-net having parameters $\theta$ and $\phi$ respectively.
The main PINN, $X$-net, predicts $\mathbf{X}=\{X_0, X_1, X_2\}$ using a single input $t$. The auxillary network, $f$-net, takes in the input from $X$-net and predicts $\mathbf{f}=\{f_1, f_2, f_3\}$ which are unknown functions. 
The loss function to be minimized can be written as $\mathcal{L}_{total} = \lambda_{data}\mathcal{L}_{data} + \lambda_{ODE}\mathcal{L}_{ODE} + \lambda_{IC}\mathcal{L}_{IC}$, where  $\lambda_{data}$, $\lambda_{ODE}$, $\lambda_{IC}$ are weights for the data, ODE and IC components of the loss function. A point worth noting is that this architecture is flexible and allows for unknown parameters to be jointly optimized with the weights and biases of the network. We found that $\lambda_{data}=1$, $\lambda_{ODE}=2$, $\lambda_{IC}=1$ yielded the best results in our experiments.
For further details on loss function, see Apx \ref{sec:app-pinns}.

\textbf{Symbolic Regression.}
For symbolic regression we consider two different methods, PySR \citep{cranmer2023_pysr} and SINDy \citep{brunton2016_sindy}. PySR uses evolutionary optimization algorithms whereas SINDy uncovers concise governing equations by selecting prominent candidate functions from a complex, high-dimensional nonlinear function space, using sparse regression techniques.
\textbf{Data generation.}
We implemented the canonical pharmacokinetic two-compartment model with first-order absorption and first-order elimination \citep{bourne2018_pk} 
\begin{equation}\label{eq-pk}
\begin{aligned}
\frac{dX_0}{dt} &= -k_a \cdot X_0 \\
\frac{dX_1}{dt} &= k_a \cdot X_0 - \left( \frac{CL + Q}{V_1} \right) \cdot X_1 + \frac{Q}{V_2} \cdot X_2 \\
\frac{dX_2}{dt} &= \frac{Q}{V_1} \cdot X_1 - \frac{Q}{V_2} \cdot X_2
\end{aligned}
\end{equation}
where \(X_0\), \(X_1\), and \(X_2\) denote the drug quantities in the depot compartment, the central compartment, and the peripheral compartment, respectively. Initially, \(X_0=1\) and both \(X_1\) and \(X_2\) are zero in all simulations. All simulations were run to $t=10$ to generate the PK datasets. 
We initialize the parameters as described in \citep{lu2014_pk_params}, for details about the parameters please see Apx. \ref{sec:app-pk-params}. We synthetically generate data sets by numerically solving Eq \eqref{eq-pk}. To simulate a real world like dataset we add Gaussian noise of different strengths to the simulated data. For this study we choose three levels of Gaussian noise i) $\mathcal{N}(0,0.005)$ ii) $\mathcal{N}(0,0.01)$ iii) $\mathcal{N}(0,0.02)$ which yields us three unique PK datasets which we refer to as low noise, medium noise and high noise respectively.

\textbf{Implementation and Training}
For $X$-net and $f$-net we choose two and three hidden layers respectively with each layer having 100 neurons. The activation function used for all layers was $\tanh$. We implemented the Adam optimizer with a learning rate of $10^{-2}$ for all models where each model was ran for 1000 epochs. We initialised the five free parameters in Eq \eqref{eq-pk} as learnable parameters within the network and gave an initial guess of unity for all parameters. We used all of the data upto $t=8$ for training the models and kept the data from $t=8-10$ for testing the quality of the model predictions. We train all of the PKINNs models using $80\%$ of the total data as training data ($t<8$) and keep the remaining $20\%$ of the data as test data ($t\geq8$) to evaluate the PKINNs in an extrapolation scenario.
Further implementation details and hyperparameter choices can be found in Apx. \ref{sec:app-pinns-training}.

\section{Results}
\begin{figure}[ht]
    \centering
\includegraphics[width=1.0\textwidth]{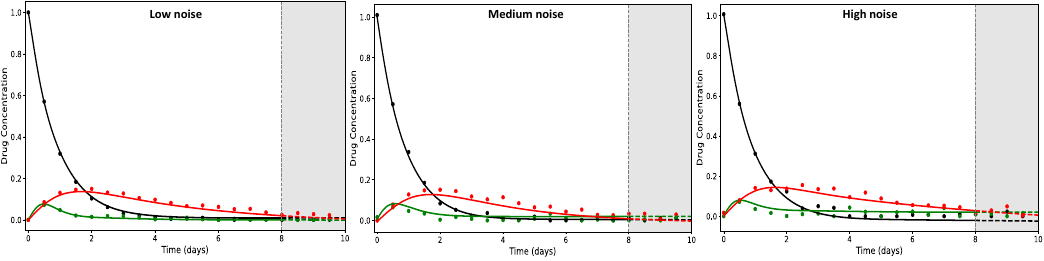}
    \caption{Drug concentration curves for the datasets. Raw data is represented as filled circles and the PKINNs model are solid lines, where the extrapolated region of the model is shown by dashed lines in the shaded region. Line colours: $X_1$ (black), $X_2$ (green), $X_3$ (red).}
    \label{fig:dc_curves}
\end{figure}

\begin{figure}[ht]
    \centering
\includegraphics[width=0.67\textwidth]{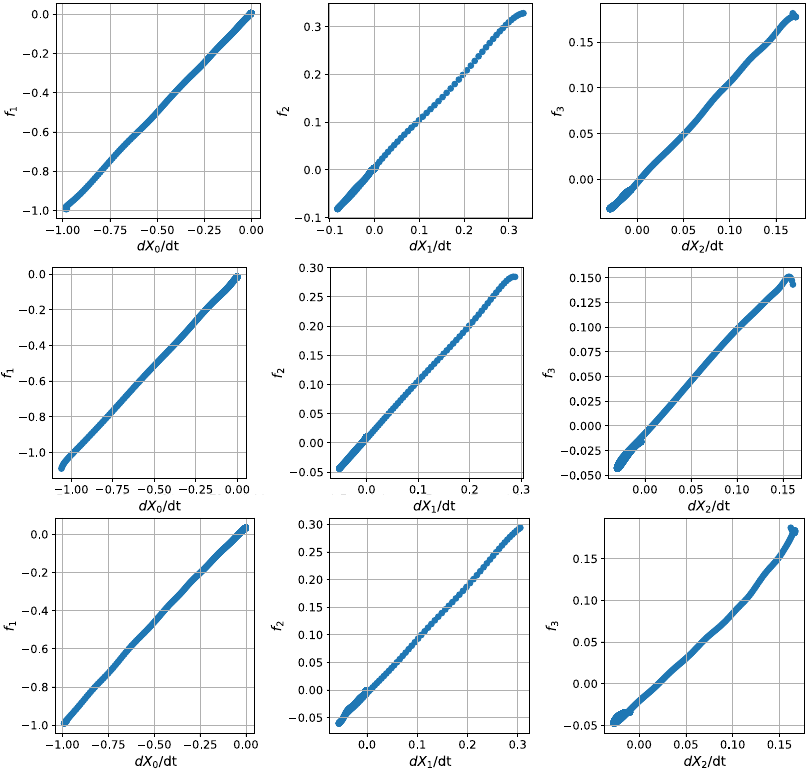}
    \caption{Comparison between the calculated derivatives and predicted derivatives of PKINNs. Low noise (upper panels), medium noise (middle panels) and high noise (upper panels).}
    \label{fig:derivatives_plot}
\end{figure}

The raw data and the model predictions for the drug concentrations are presented in Fig \ref{fig:dc_curves}. We observe that PKINNs is able to accurately model the drug concentration data, with profiles often lying on the raw data points. Interestingly, we find that PKINNs is robust to the noise in the raw data irrespective of the noise strength which is demonstrated by the ability extract the smooth drug conservation curves which are intrinsic to the multi-compartment PK model. To further probe the ability of PKINNs to capture the underlying PK model, we compare the calculated derivatives against the predicted derivatives from for each component which is presented in Fig \ref{fig:derivatives_plot}. We observe a clear linear relationship showing a very good agreement between the calculated and predicted derivatives for all drug concentration components for the low and medium noise datasets (upper and middle panels of Fig \ref{fig:derivatives_plot}). This characteristic is also observed in the profiles obtained from of the high noise dataset, though we note a slight deviation from a pure linear relationship in the third component $dX_2/dt$ vs $f_3$ (bottom right panel of Fig \ref{fig:derivatives_plot}). We find that this effect is not significant and overall PKINNs accurately models and captures the intrinsic derivatives of the underlying PK model. We have demonstrated that PKINNs performs very well in predicting the drug concentration curves for the training data, but a natural question is how well the model performs in an extrapolation setting which is important for predictions tasks in Pharmacology. \newline
The extrapolation drug concentration curves for PKINNs is represented as dashed lines in the shaded regions of Fig \ref{fig:dc_curves}. For the low and medium noise datasets we find that the extrapolations by PKINNs are faithful to the raw data with PKINNs yielding smooth curves that qualitatively match the raw data. We find that in the case of high noise dataset, the extrapolation prediction from PKINNs is not quite as good as with the lower noise models. To get a quantitative measure of the extrapolation agreement , we measure the mean squared error between the raw data and the PKINNs prediction which is given in Table \ref{tab:extrapolation_error}. From the MSE values, we see that the best agreement between the raw data and PKINNs happens for the low noise dataset \footnote{While we have shown PKINNs performs well in a extrapolation prediction scenario, we stress that the model was not designed specifically for this use case} achieving loss values of order magnitude $10^{-5}$ for $X_0$ and $X_1$ whereas the medium and high noise datasets loss values are often of order magnitude $10^{-4}$.\newline
So far we have demonstrated that PKINNs accurately predicts the drug concentration curves very well capturing the intrinsic derivatives of the datasets and is robust to the noise present in the data, however the internals of PKINNs model remain a black box. We proceed to uncover the black box to provide an explainable model yielded by PKINNs using techniques from Symbolic Regression, specifically PySR and SINDy. We apply both PySR and SINDy to the predicted drug concentration data generated by PKINNs, the results are presented in Tab \ref{tab:SR_results}. We observe that the functional forms for the drug concentration components predicted by PySR appear to be less sensitive to noise present in the data compared to the functional forms predicted by SINDy. However we find that the functional forms predicted by SINDy better match the functional form of the multi-compartment PK model used to generate the datasets which naturally leads to a more interpretable model. Additionally, as SINDy retains a more faithful functional form we can more easily interpret the values of the constants in context of the PK model compared to the functional forms yielded by PySR. We note that both models do not capture the real intrinsic ODE but we find this is not entirely surprising as noise was synthetically added to the datasets to give a more realistic data distribution, similar to what is collected from experiments.

\begin{table}[h]
\centering
\centering
\caption{Recovered expressions for the approximate ODE components $\Tilde{f1}$, $\Tilde{f2}$ and $\Tilde{f3}$ from PKINNs using both PySR and SINDy. Each row corresponds to the results for different levels of noise present in the raw data (low, medium and high). }
\resizebox{\columnwidth}{!}{
\begin{tabular}{ccccccc}
\toprule
\multicolumn{1}{c}{} & \multicolumn{2}{c}{\textbf{PySR}} &  \multicolumn{4}{c}{\textbf{SINDy}} \\
\cmidrule(rl){1-4} \cmidrule(rl){4-7}
\multicolumn{1}{c|}{\textbf{Noise}}  & \multicolumn{1}{c|}{\(\Tilde{f1}\)} & \multicolumn{1}{c|}{\(\Tilde{f2}\)} & \multicolumn{1}{c|}{\(\Tilde{f3}\)} & \multicolumn{1}{c|}{\(\Tilde{f1}\)} & \multicolumn{1}{c|}{\(\Tilde{f2}\)} & \(\Tilde{f3}\)\\
\midrule
\multicolumn{1}{c|}{Low} & \multicolumn{1}{c|}{\(-1.1X_0\)} & \multicolumn{1}{c|}{\((-0.6+X_0)X_0\)} & \multicolumn{1}{c|}{\(-1.1X_0+0.2X_2\)} & \multicolumn{1}{c|}{\(-X_0-0.8X_1\)} & \multicolumn{1}{c|}{\(0.4X_0-3.4X_1-0.2X_2\)}& \(2.3X_1-0.3X_2\) \\ 
\multicolumn{1}{c|}{Medium}  & \multicolumn{1}{c|}{\(-1.2X_0\)} &\multicolumn{1}{c|}{\((-0.6+X_0)X_0\)} &\multicolumn{1}{c|}{\(0.2X_0\)} &\multicolumn{1}{c|}{\(-1.2X_0\)}  & \multicolumn{1}{c|}{\(0.4X_0-2.8X_1\)}&\(2.2X_1-0.5X_2\)  \\
\multicolumn{1}{c|}{High}  & \multicolumn{1}{c|}{\(-1.2X_0\)} & \multicolumn{1}{c|}{\((-0.6+X_0)X_0+X_1\)} &\multicolumn{1}{c|}{\(0.2X_0\)}  & \multicolumn{1}{c|}{\(-1.1X_0-0.4X_1-0.3X_2\)} & \multicolumn{1}{c|}{\(0.2X_0-1.3X_1-0.3X_2\)} &\(0.2X_0-0.2X_2\)  \\
\bottomrule
\end{tabular}}
\label{tab:SR_results}
\end{table}

\section{Impact statement}
In this work we have developed a novel pharmacokinetic informed neural network model called PKINNs, which is used to model drug concentration data yielded from pharmacokinetic experiments. We demonstrate that PKINNs accurately and robustly predicts the intrinsic derivatives of the underlying PK model and performs well in extrapolation predictions scenarios. We show that the dynamical models yielded by PKINNs are interpretable and explainable using techniques from SR, namely PySR and SINDy, and find the functional forms from SINDy to well predict the intrinsic PK model.  Our goal is to enhance PKINNs to handle arbitrary compartment-based PKPD and complex quantitative systems pharmacology (QSP) models. This extension is particularly impactful as manually deriving such models is labor-intensive. The data-driven approach presented here explores function space, offering various parsimonious models for practitioners to choose from. Even if an expert disagrees with the suggested system, the framework serves as a valuable starting point for further derivation and selection in model-informed drug discovery.

\newpage
\bibliographystyle{plain}
\bibliography{bibliography}

\appendix

\section{PKINNs details}\label{sec:app-pinns}
\subsection{Loss function}\label{sec:app-pinns-loss}

Given a model that predicts $\mathbf{X}$ and $\mathbf{f}$ for a temporal coordinate $t$, where $\mathbf{X}$ is a vector function with three components $X_0, X_1, X_2$, the residuals for each of these components are calculated as follows:
\begin{align*}
\text{res}_1 &= \frac{dX_0}{dt} - f_1, \\
\text{res}_2 &= \frac{dX_1}{dt} - f_2, \\
\text{res}_3 &= \frac{dX_2}{dt} - f_3,
\end{align*}
where $f_i$ are the predicted ODE components from PKINNs.
These residuals represent the difference between the derivatives of the model predictions and the predicted values. The ODE loss function can be formulated as the mean square of these residuals:
\begin{equation}
\mathcal{L}_{ODE} = \frac{1}{N} \sum_{i=1}^{N_{ODE}} \left( \sum_{i=1}^{N_c} \left[ \frac{dX_{i-1}}{dt} - f_i \right] \right)^2
\end{equation}
where $N_{ODE}$ is the number of collocation points enforcing the ODE residual loss and $N_c$ are the number of compartments of the PK model.\newline
The loss function of the initial condition is defined as the error between a constant initial condition vector and the predicted values, it can be represented as:
\begin{equation}
\mathcal{L}_{IC} = \sqrt{\sum_{i=0}^{N_c} (X_i - X'_i)^2}.
\end{equation}

The final component of the loss functions is the data loss term, defined as the mean squared error between the true values \( \mathbf{X} \) and the predicted values \( \mathbf{X'} \) from the model:

\begin{equation}
\mathcal{L}_{data} = \frac{1}{N} \sum_{i=0}^{N_{data}} \left(X_i - X' \right)^2.
\end{equation}
The total loss function combines all the separate components with component specific weighting, which can be written as:

\begin{equation}
\mathcal{L}_{total} = \lambda_{data}\mathcal{L}_{data} + \lambda_{ODE}\mathcal{L}_{ODE} + \lambda_{IC}\mathcal{L}_{IC}.
\end{equation}

Throughout our numerical experiments, we empirically found that $\lambda_{data}=1$, $\lambda_{ODE}=2$, $\lambda_{IC}=1$ yielded the best performance for PKINNs.

\subsection{Model training details}\label{sec:app-pinns-training}
All of the deep learning models were implemented using \textsc{Keras} in \textsc{TensorFlow}. The input dimension into $X$-net was one and the output dimension was three, to match the number of variables within the PK model studied. The input dimensions into $f$-net was set to four and the output dimension was set to three to match the number of ODEs of the PK model. We experimented with a number of learning rates using the Adam optimizer but we found that $10^{-2}$ yielded us the best performance. We also found that 1000 epochs was sufficient to achieve convergence of the PKINNs model for the experiments we ran. Interestingly, we found a notable improvement in the model performance when we set $\lambda_{ODE}=2$ from $\lambda_{ODE}=1$.

\subsection{Extrapolation Error}
We measure the mean squared error of the raw data against the predictions made by PKINNs. The results are given in in Table \ref{tab:extrapolation_error}.
\begin{table}[h]
\centering
\caption{MSE of the PKINNs extrapolation prediction against the raw data for the different drug concentration components $X_0$, $X_1$ and $X_2$.}
\begin{tabular}{|c|c|c|c|}
\hline
Noise Level & MSE $X_0$ & MSE $X_1$ & MSE $X_2$ \\ \hline
Low         &   $6.2 \times 10^{-5}$   &  $1.9 \times 10^{-5}$      &   $1.6 \times 10^{-4}$     \\ \hline
Medium      &   $3.1 \times 10^{-5}$     &  $2.6 \times 10^{-4}$      &  $5.7 \times 10^{-4}$       \\ \hline
High        & $7.1 \times 10^{-4}$   &   $1.7 \times 10^{-4}$      &  $5.3 \times 10^{-4}$      \\ \hline
\end{tabular}
\label{tab:extrapolation_error}
\end{table}

\section{Pharmacology model details}\label{sec:app-pk}
\subsection{Parameters}\label{sec:app-pk-params}
The model parameters are as follows: \(\mathrm{k_a}\)=1.14 represents the absorption rate constant; \(\mathrm{CL}\)=3.57 is the elimination clearance rate; \(\mathrm{Q}\)=1.14 denotes the inter-compartment distribution; and \(V_1\)=0.454, \(V_2\)=2.87 are the central and peripheral volumes of distribution, respectively.

\end{document}